\documentclass[runningheads]{llncs}
\usepackage[T1]{fontenc}
\usepackage{graphicx}
\usepackage{hyperref}
\usepackage{color}

\usepackage[colorinlistoftodos]{todonotes}
\usepackage{caption}
\usepackage{subcaption}
\usepackage{amsmath}
\usepackage{amsfonts}
\usepackage{flushend}

\usepackage{graphicx}
\usepackage{booktabs}  
\usepackage{longtable} 
\usepackage{multirow} 
\usepackage{svg}
\usepackage{multirow}
\usepackage{tabularx}

\usepackage[capitalize,nameinlink]{cleveref}
\crefformat{figure}{#2\textup{Fig.\,#1}#3}
\Crefformat{figure}{#2\textup{Fig.\,#1}#3}
\crefname{figure}{Fig.}{Figs.}
\Crefname{figure}{Fig.}{Figs.}

\begin{document}
\title{A Framework for Deploying Learning-based Quadruped Loco-Manipulation}

\makeatletter
\newcommand{\printfnsymbol}[1]{%
  \textsuperscript{\@fnsymbol{#1}}%
}
\makeatother

\author{Yadong Liu \thanks{These authors contributed equally.}\inst{1} \and
        Jianwei Liu \printfnsymbol{1}\inst{1} \and
        He Liang\inst{2} \and
        Dimitrios Kanoulas\inst{1}}

\authorrunning{Y. Liu et al.}
\titlerunning{A Framework for Deploying Learning-based Quadruped Loco-Manipulation}

\institute{Department of Computer Science, University College London, Gower Street, London, WC1E 6BT, UK.\\
\email{\{yadong.liu.20, jianwei.liu.21, d.kanoulas\}@ucl.ac.uk} 
 \and Department of Computer Science, University of Oxford, Parks Rd, Oxford, OX1 3QG, UK\\
\email{he.liang@cs.ox.ac.uk}
}
\maketitle              
\vspace{-20pt}
\begin{abstract}
Quadruped mobile manipulators offer strong potential for agile loco-manipulation but remain difficult to control and transfer reliably from simulation to reality. Reinforcement learning (RL) shows promise for whole-body control, yet most frameworks are proprietary and hard to reproduce on real hardware. We present an open pipeline for training, benchmarking, and deploying RL-based controllers on the Unitree B1 quadruped with a Z1 arm. The framework unifies sim-to-sim and sim-to-real transfer through ROS, re-implementing a policy trained in Isaac Gym, extending it to MuJoCo via a hardware abstraction layer, and deploying the same controller on physical hardware. Sim-to-sim experiments expose discrepancies between Isaac Gym and MuJoCo contact models that influence policy behavior, while real-world teleoperated object-picking trials show that coordinated whole-body control extends reach and improves manipulation over floating-base baselines. The pipeline provides a transparent, reproducible foundation for developing and analyzing RL-based loco-manipulation controllers and will be released open source to support future research.
\end{abstract}

\vspace{-20pt}


\section{Introduction}
\vspace{-10pt}

\begin{figure}[h!]
    \centering
    \includegraphics[width=0.9\linewidth]{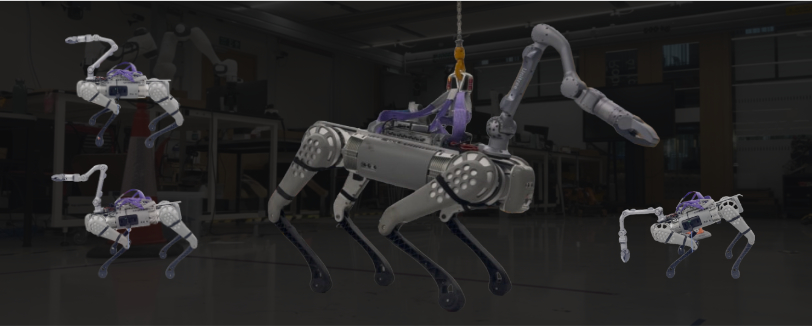}
    \caption{Wholebody control motion deployed on Real-world robot hardware}
    \label{fig:banner}
    \vspace{-15pt}
\end{figure}

Legged mobile manipulators combine the terrain adaptability of quadrupeds with the dexterity of robotic arms, enabling tasks such as stair climbing~\cite{portela2025whole}, rough terrain traversal~\cite{jenelten2024dtc}, and disaster response~\cite{klamtsupervised}. However, these platforms are inherently high-dimensional (e.g., Unitree B1+Z1 has 19 DoF), making effective coordination between locomotion and manipulation a longstanding challenge~\cite{liu2024visual,hou2025efficient}. Most existing systems decouple locomotion and manipulation, simplifying control but often limiting coordination and generalization. In contrast, \emph{whole-body control} (WBC) seeks to unify all degrees of freedom, allowing legs to extend the arm’s workspace and arms to contribute to balance~\cite{fu2023deep,liu2024visual,hou2025efficient,zhu2025versatile,wang2025quadwbg}.

Low-level WBC strategies can be broadly categorized into three classes. Optimization-based approaches (e.g. MPC, NMPC)~\cite{sleiman2021unified,zimmermann2021go} produce interpretable multi-contact behaviors but require accurate models and significant computation. Sampling-based methods (e.g. RRT, contact planning) address hybrid contacts but remain task-specific and computationally heavy~\cite{dalibard2010manipulation}. More recently, reinforcement learning (RL) has emerged as a compelling alternative, producing unified policies that achieve smoother coordination, extended reach, and robust recovery from disturbances~\cite{fu2023deep,liu2024visual,zhi2025learning}. Yet, despite rapid progress, RL-based loco-manipulation frameworks remain largely proprietary and hardware-specific, limiting reproducibility and systematic evaluation. In particular, no open end-to-end RL pipeline currently supports whole-body control or sim-to-real deployment on the Unitree B1+Z1, a widely used quadruped-manipulator platform.

\textbf{In this work}, we address this gap by developing a transparent and extensible pipeline for whole-body RL research. Our framework supports reproducible training, sim-to-sim benchmarking, and deployment on real hardware. While the codebase is not yet public, we intend to release it as open-source in the near future, providing a foundation for advancing loco-manipulation towards robust, real-world applications.
\vspace{-5pt}

\section{Related works}\label{sec:rw}
\vspace{-10pt}

Recent reinforcement learning (RL) approaches to whole-body control (WBC) span monolithic, hierarchical, adaptive, and hybrid designs. \textbf{Monolithic policies} (e.g., Deep WBC~\cite{fu2023deep}, Unified Force Position~\cite{zhi2025learning}, Efficient Unified Policy~\cite{hou2025efficient}) integrate locomotion and manipulation within a single controller, exploiting inter-limb coupling for seamless coordination. These methods leverage advantage mixing, proprioceptive force inference, and structural priors, though scaling to high-dimensional platforms makes training challenging. \textbf{Hierarchical frameworks} decompose control into high-level planning and low-level execution. Representative examples include terrain-conditioned skill libraries (ANYmal Parkour~\cite{hoeller2024anymal}), adaptive limb allocation (ReLIC~\cite{zhu2025versatile}), perception-informed stacks~\cite{liu2024visual,stamatopoulou2024dippest,liu2024dipper,liu2023vit}, bi-level diffusion–RL systems (UMI-on-Legs~\cite{ha2024umi}), and recent integrated loco-manipulation frameworks~\cite{zhou2022teleman}. These improve modularity and transfer, but depend on carefully designed task interfaces and rewards. \textbf{Adaptive and recurrent architectures} enhance robustness by estimating latent extrinsics (e.g., RMA~\cite{kumar2021rma}, ROA~\cite{fu2023deep}) or by introducing temporal reasoning with GRU/LSTM modules. While effective, these often suffer from instability in high-DoF settings~\cite{yao2024local,beddow2024reinforcement}.

Observation and action spaces typically follow a compact proprioceptive + joint-target template, with PD stabilization. Recent extensions incorporate vision for context~\cite{agarwal2023legged,liu2024visual,hadjivelichkov2024reinforcement}, hybrid command spaces combining base velocities and deltas of the end-effector, and estimation of proprioceptive force for compliant interaction. Reward design extends locomotion objectives with manipulation accuracy, perception alignment, and curriculum-based progression~\cite{ellis2023navigation}.

In general, representative systems illustrate complementary strengths: VBC embodies perception–control hierarchies; Deep WBC and Unified Force–Position highlight unified adaptation; ReLIC and UMI-on-Legs demonstrate flexible coordination; and QuadWBG integrates RL locomotion with grasp planning. Despite architectural diversity, most pipelines converge on a common substrate—compact proprioceptive observations, joint-target actions with PD stabilization, and PPO optimization—differentiated primarily by how perception, adaptation, and planning modules are composed. Notably, while some code bases are partially open-sourced, most omit real-hardware deployment stacks, hindering rigorous replication and benchmarking.
\vspace{-10pt}

\section{Method}\label{sec:method}
\begin{figure}[ht!]
\vspace{-15pt}
  \centering
  \includegraphics[width=0.9\linewidth]{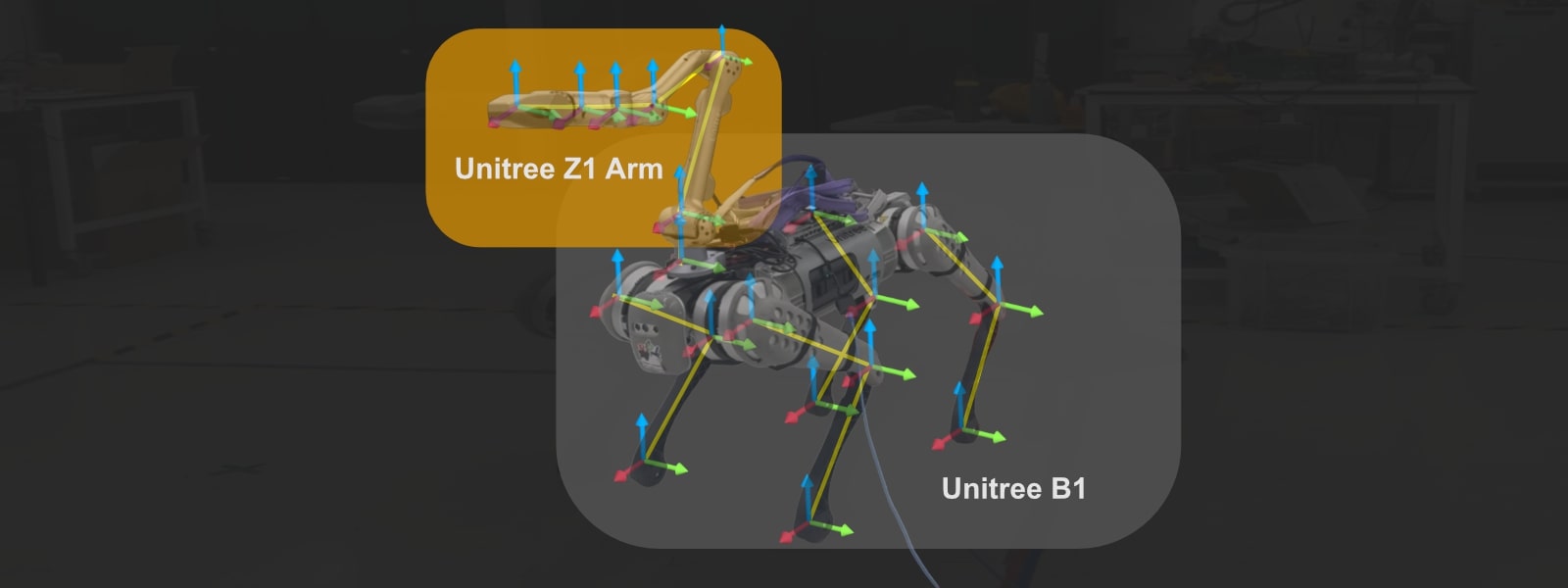}
  \caption{Unitree B1 and Z1 platform setup used for development of the framework.}
  \label{fig:b1z1_platform}
  \vspace{-15pt}
\end{figure}

In this work, we build our pipeline on the Unitree B1 quadruped~\cite{unitreeB1} equipped with a Unitree Z1 arm mounted on torso \cite{unitreeZ1}, as shown in \cref{fig:b1z1_platform}. The hardware platform provides 19 degrees of freedom (DoFs): the B1 contributes 12 actuated joints (hip, thigh, and knee for each leg), while the Z1 adds 6 arm DoFs and a 1-DoF gripper. Our setup is as follows~\cite{liu2024visual}. 

\subsection{Low-Level Policy Architecture and Training}

We adopt the partially open-source framework of Liu \textit{et al.}~\cite{liu2024visual} as our baseline, extending it to our framework pipeline then deployed on the Unitree B1+Z1 platform.~\cite{makoviychuk2021isaac}. 

\subsubsection{RL formulation.}  
The controller is trained as a reinforcement learning policy in a Markov Decision Process (MDP),
\[
  \mathcal{M} = \langle \mathcal{S}, \mathcal{A}, \mathcal{P}, r, \gamma, \rho_0 \rangle,
\]
with state space $\mathcal{S}$ (proprioception, joint states, contacts), actions $\mathcal{A}$ (joint targets), transition dynamics $\mathcal{P}$, reward $r$, discount $\gamma$, and initial distribution $\rho_0$.  
At each step $t$, the policy $\pi_\theta$ produces $a_t \sim \pi_\theta(\cdot|s_t)$, maximizing the return
\[
  J(\pi_\theta) = \mathbb{E}\!\left[\sum_{t=0}^\infty \gamma^t r(s_t,a_t)\right].
\]
Partial observability is addressed through observation histories, recurrent encoders, and domain randomization.  

\subsubsection{Policy architecture.}
\begin{figure}[ht!]
    \centering
    \includegraphics[width=0.95\linewidth]{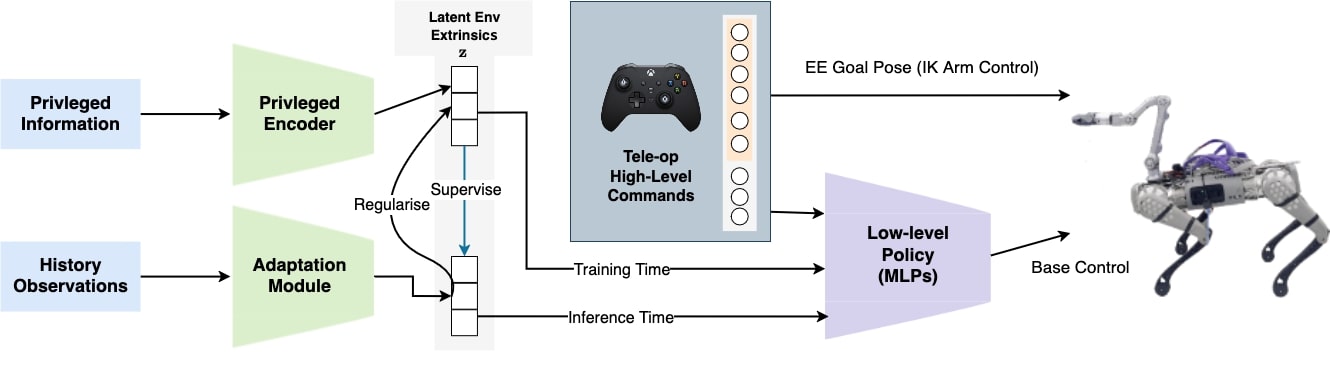}
    \caption[Policy network architecture]{Inputs (proprioception, previous action, phase cues, command, extrinsics)
    pass through a 3-layer MLP to produce 12 leg joint targets. A privileged encoder (training) and an adaptation module (deployment) provide $z_t$ for dynamics variation.}
    \label{fig:policy_arch}
    \vspace{-15pt}
\end{figure}

The low-level policy $\pi_\theta$ is a 3-layer MLP (128 units, ELU) with a $\tanh$ output head for 12 leg joint targets (\cref{fig:policy_arch}). The inputs concatenate proprioception, previous action, gait phase cues, high-level command $c_t$, and a latent extrinsic vector $z_t$. The latter is supervised with privileged parameters during training and inferred online at deployment via an ROA-style adaptation module. Arm control is analytically handled by damped least-squares IK, following~\cite{liu2024visual}, leaving RL to focus on coordinated body behaviors.

The command input $c_t = (v^{\mathrm{cmd}}, \omega^{\mathrm{cmd}}, \mathbf{p}^{\mathrm{ee,cmd}}, \mathbf{R}^{\mathrm{ee,cmd}})$ specifies the base velocity and the end-effector pose goals. The policy outputs normalized leg joint positions, mapped to admissible ranges, and tracked by the onboard PD controllers.

\subsubsection{Optimization.}
Policies are trained with Proximal Policy Optimization (PPO)~\cite{schulman2017proximal}, whose cut surrogate objective is
\[
  L_{\mathrm{CLIP}}(\theta)=\mathbb{E}_t\!\left[\min\big(\rho_t(\theta)\hat{A}_t,\ \mathrm{clip}(\rho_t(\theta),1-\epsilon,1+\epsilon)\hat{A}_t\big)\right],\quad
  \rho_t(\theta)=\tfrac{\pi_\theta(a_t|s_t)}{\pi_{\theta_{\mathrm{old}}}(a_t|s_t)}.
\]
where $\rho_t(\theta)=\tfrac{\pi_\theta(a_t|s_t)}{\pi_{\theta_{\mathrm{old}}}(a_t|s_t)}$ is the probability ratio between current and old policies, 
$\hat{A}_t$ is the estimated advantage, and $\epsilon$ is the clipping range. 

We combine this with a value loss and entropy regularization to encourage exploration and estimate advantages using Generalized Advantage Estimation (GAE)~\cite{schulman2015high}. Training runs in Isaac~Gym with thousands of parallel environments, enabling large-batch on-policy updates.  

\subsubsection{Training setup.}  
Rewards combine base-velocity tracking, energy efficiency, alive stability, and gait regularization, with exponential kernels stabilizing velocity tracking. Commands $c_t=(v^{\mathrm{cmd}},\omega^{\mathrm{cmd}},\mathbf{p}^{\mathrm{ee}},\mathbf{R}^{\mathrm{ee}})$ are sampled in a wide range of velocities and poses of the end-effector. To help bridge the sim-to-real gap, we apply domain randomization over terrain friction and geometry, robot parameters (mass, CoM (Centre of Mass), damping, latency) and sensor noise. Curriculum learning gradually increases the difficulty of randomization, while extrinsics latent $z_t$ enables online adaptation to unmodeled dynamics. This design yields a transferable low-level controller capable of whole-body loco-manipulation on the B1+Z1 platform.

\subsection{Deployment}
\begin{figure}[htbp]
    \centering
    \vspace{-15pt}
     \includegraphics[width=0.95\linewidth]{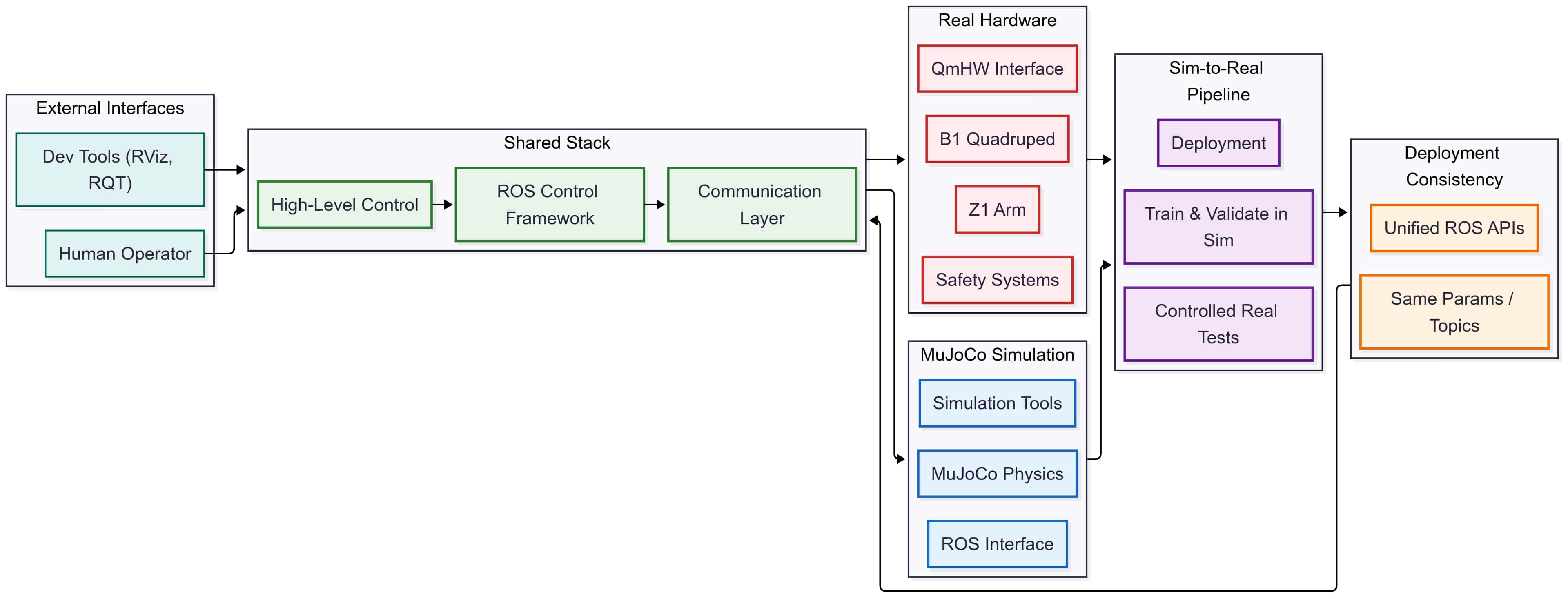}
    \caption[Deployment architecture overview]{Deployment architecture of the proposed system. The same policy and controller stack run across both real and simulated environments, with only the hardware interface replaced.}
    \label{fig:deployment_arch_overall}
    \vspace{-15pt}
\end{figure}

As a core contribution of this work, our unified deployment pipeline supports both sim-to-sim verification and real-world deployment on the Unitree B1 quadruped with a Z1 manipulator. The framework is implemented in the Robot Operating System (ROS)~\cite{quigley2009ros}, with \texttt{ros\_control} to ensure thread safety and real-time low-level control performance. The only difference between simulation and hardware execution lies in the interface layer, as illustrated in \cref{fig:deployment_arch_overall}.

The framework is organized into five layers. At the top, \textbf{the consistency} of the deployment enforces identical nodes, parameters, and variant definitions across the simulation and hardware. \textbf{External interfaces} provide teleoperation and access to development tools such as \texttt{RViz}, \texttt{Foxglove}, and \texttt{RQT Dynamic Reconfigure} for debugging and real-time tuning. The core \textbf{shared software stack} contains the policy, finite state machine, goal management, and unified PD controllers, all invariant across environments. The execution then branches into either the \textbf{simulation environment}, which integrates MuJoCo physics, a \texttt{MujocoHW} emulation layer, and data collection utilities, or the \textbf{real hardware environment}, which connects via the \texttt{HW} interface to the B1, Z1 and associated safety systems. This layered design ensures seamless transfer from simulation to real-world deployment without modifications to the policy or control stack.

\subsubsection{Sim2sim}
\begin{figure}[htbp]
    \centering
    \includegraphics[width=1\linewidth]{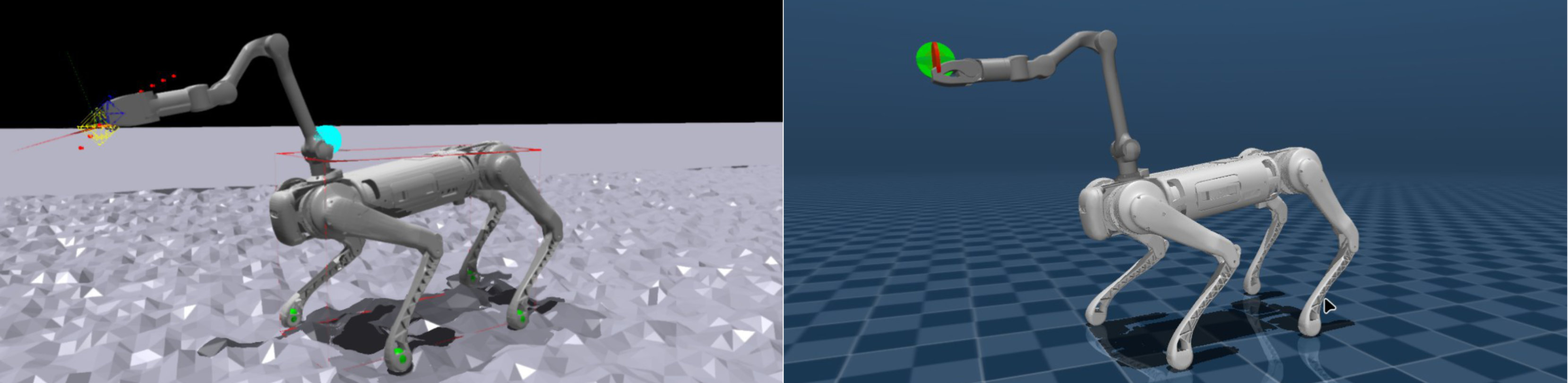}
    \caption{Simulation environments used for sim2sim loco-manipulation experiments: Isaac Gym (Left) and Mujoco (Right).}
    \label{fig:sim_envs}%
    \vspace{-15pt}
\end{figure}

Since there is no publicly available sim-to-sim pipeline for the Unitree B1+Z1 platform, we reconstructed the robot assets from its Unified Robot Description Format (URDF) and generated the corresponding MuJoCo scenes (\cref{fig:sim_envs}).

\begin{figure}[htbp]
    \centering
    \includegraphics[width=0.9\linewidth]{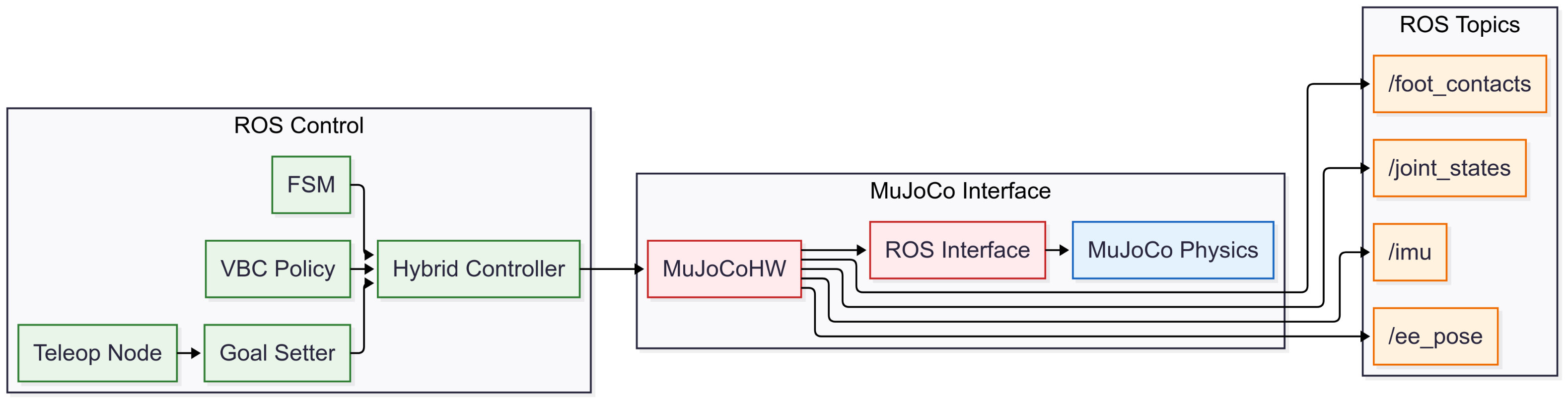}
    \caption[MuJoCo hardware interface architecture]{Architecture of the MuJoCo hardware interface. The interface emulates the same \texttt{RobotHW} abstraction as on real hardware, ensuring compatibility of the control stack.}
    \label{fig:mujoco_ros_hw}
    \vspace{-15pt}
\end{figure}

To unify simulation and deployment, we developed a \texttt{MujocoHW} interface that emulates the \texttt{hardware\_interface::RobotHW} abstraction used on the physical platform. This enables the same policy and controller stack to operate without modification across domains (\cref{fig:mujoco_ros_hw}).

The interface consists of two layers. At the low level, the MuJoCo engine simulates rigid-body dynamics and publishes raw sensor streams (e.g., \texttt{/mujoco\_sim/\allowbreak joint\_states}, \texttt{/imu\_data}, \texttt{/foot\_contacts}). At the higher level, \texttt{MujocoHW} maps these into standard ROS control interfaces (\texttt{JointStateInterface}, \texttt{Hybrid\allowbreak JointInterface}, \texttt{ImuSensorInterface}, \texttt{ContactSensorInterface}). Controller outputs are translated back into MuJoCo commands (\texttt{/legs\_cmd}, \texttt{/z1/arm\_cmd}), completing the feedback loop. From the controller’s perspective, the simulated pipeline is indistinguishable from hardware.

The framework also supports a B1-only variant via a lightweight “fake arm” module, which generates synthetic arm states through inverse kinematics. This preserves interface consistency while avoiding risks to real hardware.

Finally, the interface preserves the observation–action contract from Isaac Gym. Simulation physics runs at 200\,Hz and the ROS control loop at 500\,Hz to match real-robot timing. Observations (base state, joint states, contact indicators, previous actions, end-effector goals) are indexed identically to training. Leg commands are executed via PD control, while manipulator goals are tracked through inverse kinematics. Contact events are derived from simulated forces and threshold for binary semantics consistent with training.

\subsubsection{Sim-to-Real Deployment}
\begin{figure}[htbp]
    \centering
    \vspace{-15pt}
    \includegraphics[width=0.7\linewidth]{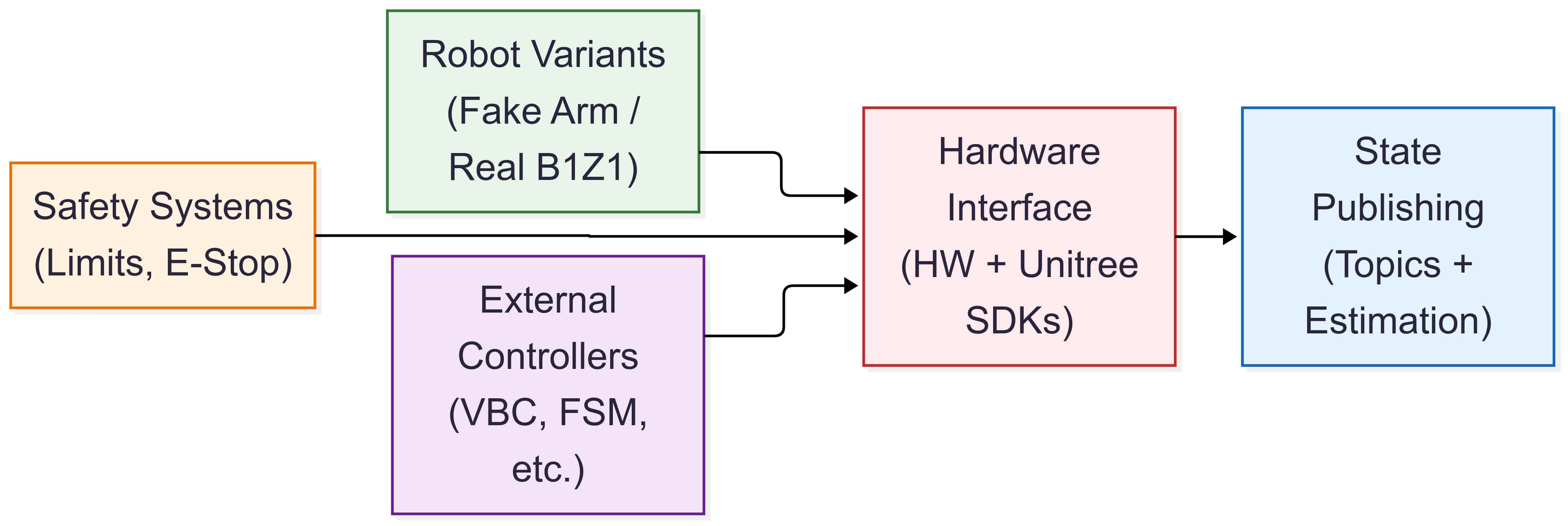}
    \caption[Real robot hardware interface architecture]{Real robot hardware interface (\texttt{HW}), integrating Unitree SDKs into ROS control with sensing, safety, and variant support.}
    \label{fig:real_robot_hw}
    \vspace{-15pt}
\end{figure}

On hardware, the control stack interfaces with the Unitree B1+Z1 via the \texttt{QmHW} module, implemented in C++ on \texttt{ros\_control}. Similar to the \texttt{MujocoHW} emulator, it bridges ROS controllers with Unitree SDKs over UDP, exposing standard abstractions (\texttt{JointStateInterface}, \texttt{HybridJoint\allowbreak Interface}, \texttt{ImuSensorInterface}, \texttt{ContactSensorInterface}) at 500\,Hz (\cref{fig:real_robot_hw}).

\paragraph{Foot Contact Estimation.}
Since B1 lacks foot sensors, we estimate contacts from joint torques $\boldsymbol{\tau}$ and dynamics using $\mathbf{F}_{\text{contact}} = (\mathbf{J}^T)^{-1}\boldsymbol{\tau}$. The per-leg contact forces $\mathbf{F}_{\text{contact}}$ are computed using the SVD-based pseudo-inverse of Jacobian $\mathbf{J}$, with clamping at 1000 \,N to reject spikes. A binary contact flag is set if $|F_z| > 80$\,N, consistent with the load specs of B1. This runs at 500\,Hz and provides stable stance detection without additional sensors.

\subsubsection{Teleop and visualisation}
\begin{figure}[htbp]
    \centering
    \vspace{-15pt}
    \includegraphics[width=1\linewidth]{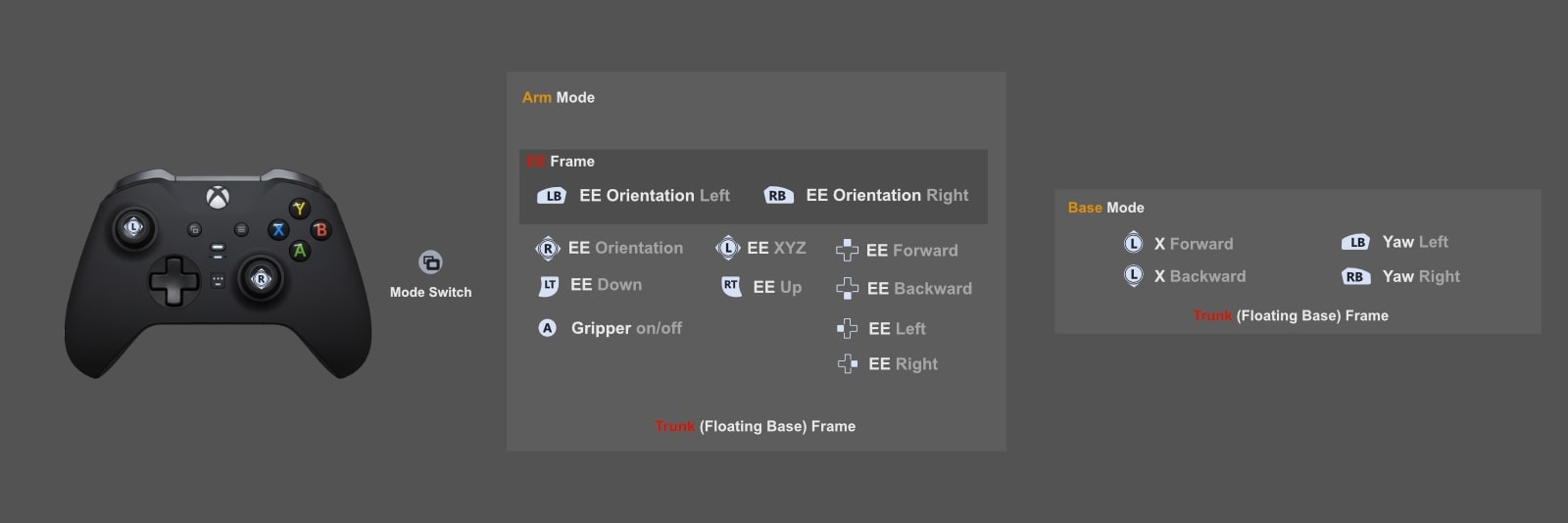}
    \caption[Joystick teleoperation interface]{Dual-mode joystick teleoperation for base locomotion (left stick/triggers) and arm manipulation (right stick/buttons).}
    \label{fig:teleop_interface}
    \vspace{-15pt}
\end{figure}

The interface supports dual-mode joystick control (\cref{fig:teleop_interface}). In \textbf{base mode}, inputs map to body velocity commands ($v_x, v_y, v_{yaw}$); in \textbf{arm mode}, they incrementally update the end-effector pose in the trunk frame. End-effector goals are persistent, updated smoothly to avoid abrupt motions, and constrained within workspace bounds. Inputs pass through deadzones, scaling curves, and rate limits for safe and responsive control.

\begin{figure}[htbp]
    \centering
    \vspace{-15pt}
    \includegraphics[width=0.9\linewidth]{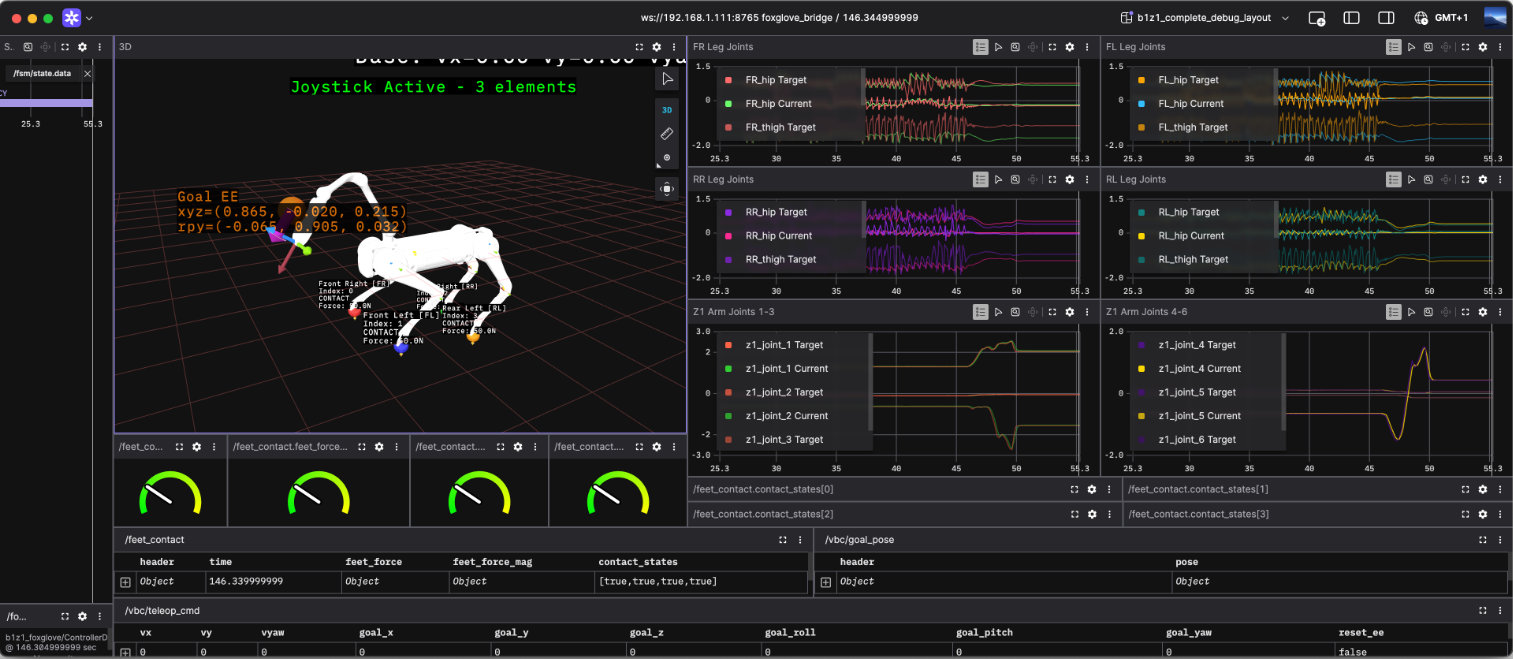}
    \caption[Foxglove visualisation]{Foxglove visualisation of loco-manipulation with real-time joint, contact, and policy feedback.}
    \label{fig:foxglove_viz}
    \vspace{-15pt}
\end{figure}

For monitoring, the framework integrates with Foxglove Studio (\cref{fig:foxglove_viz}), which streams downsampled topics such as joint states, IMU data, contact events, and end-effector goals. Custom panels visualize force vectors, joystick inputs, and policy execution, while adaptive filtering and compression ensure responsiveness without affecting the 500\,Hz control loop.

\vspace{-5pt}
\section{Experiments}\label{sec:exp}
We evaluate the learned whole-body policy in three settings: (i) \emph{sim-to-sim} tracking across Isaac~Gym and MuJoCo, (ii) \emph{sim-to-real} tracking on the Unitree B1+Z1 platform, and (iii) \emph{teleoperated object picking}. Across all experiments, the same ROS stack and policy are used; only the hardware interface differs. We use fixed end-effector (EE) poses and predefined base-velocity commands, reporting RMSE between commanded and measured velocities to quantify tracking, and we assess stability, reachability, and task success for manipulation. We summarise protocols and metrics below, then present results for each setting.

\vspace{-10pt}
\subsection{Sim-to-sim Tracking}

\begin{figure}[ht!]
    \centering
    \vspace{-15pt}
    \includegraphics[width=0.7\linewidth]{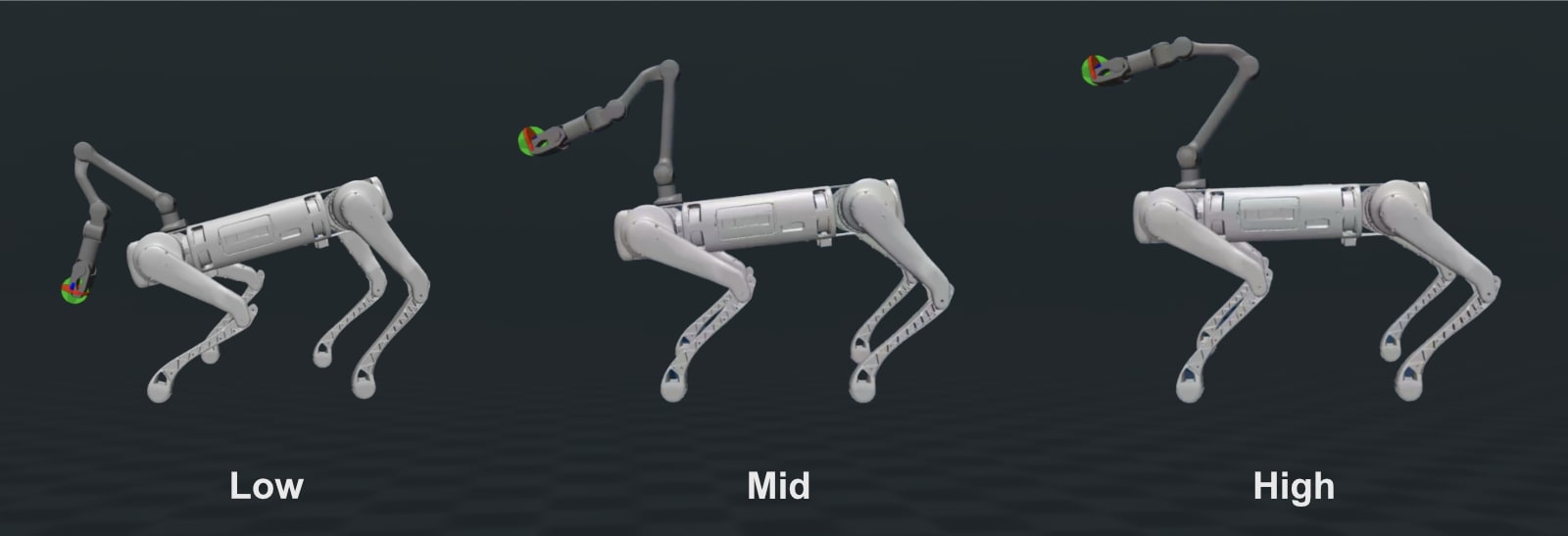}
    \caption[Sim-to-sim experiment EE goals]{Selected end-effector goals for sim-to-sim evaluation. Three targets were chosen to span low, mid (neutral), and high body configurations.}
    \label{fig:sim2sim_exp_goal_ee_mujoco}
    \vspace{-15pt}
\end{figure}

We evaluated whether the same policy yields consistent behavior across simulators under identical inputs. 
Three end-effector (EE) poses were fixed—low, mid/neutral, and high (\cref{fig:sim2sim_exp_goal_ee_mujoco}). Then, a 23\,s sequence of forward, backward, turning, and combined motions was executed. The tracking accuracy was measured as the root-mean-square error (RMSE) between the commanded and measured base velocities, 
$\text{RMSE}(v_d) = \sqrt{\tfrac{1}{N}\sum_{t=1}^N (v_d^{\text{meas}}(t) - v_d^{\text{cmd}}(t))^2}$, with $d \in \{x,\text{yaw}\}$.

\subsubsection{Isaac~Gym.} 
\begin{figure}[htbp!]
    \centering
    \includegraphics[width=0.75\linewidth]{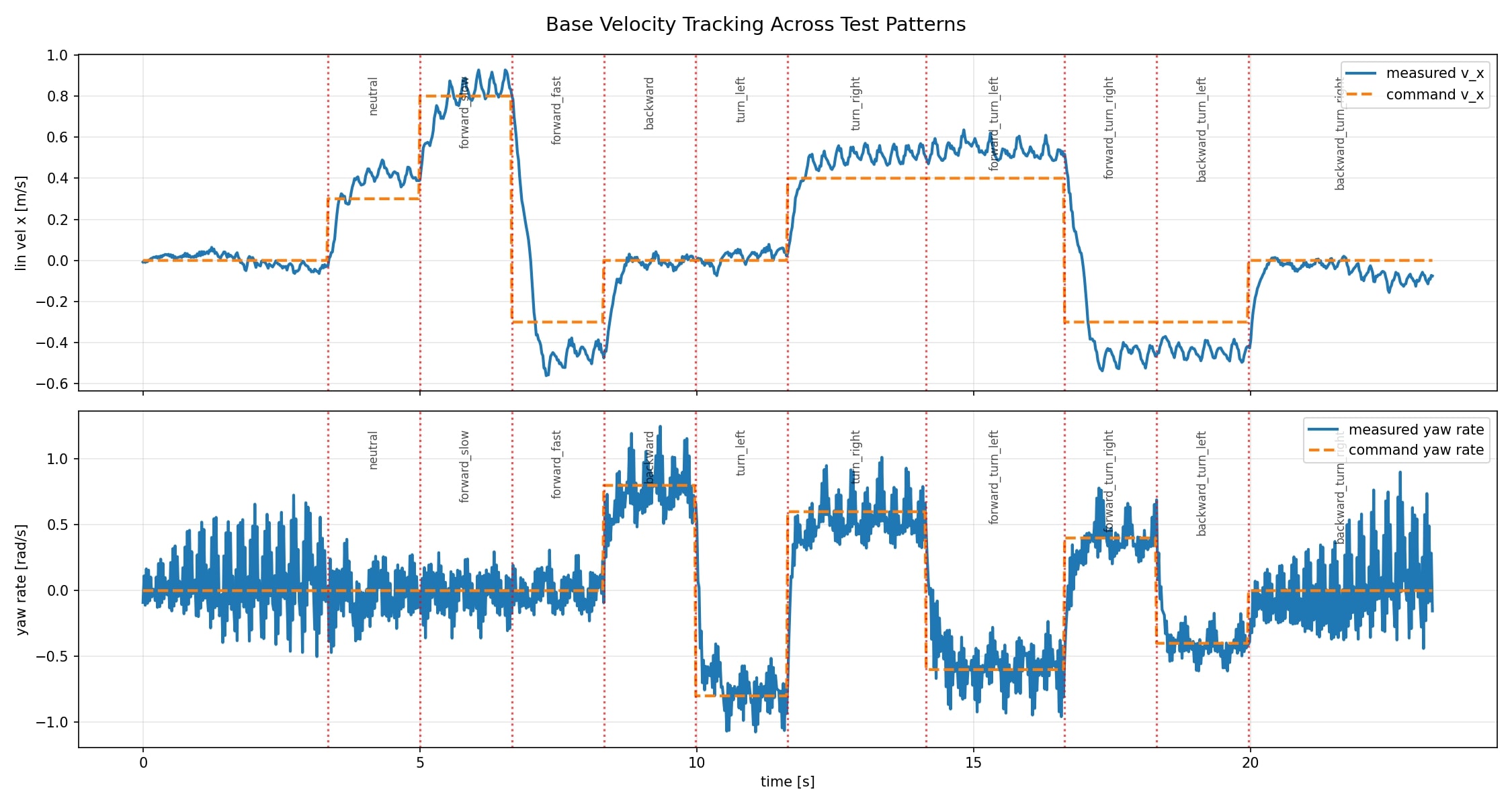}
    \caption{Base velocity tracking with low end-effector configuration in Isaac~Gym}
    \label{fig:base_velocity_tracking_isaac_low}
    \vspace{-15pt}
\end{figure}

\begin{figure}[htbp]
    \centering
    \includegraphics[width=0.8\linewidth]{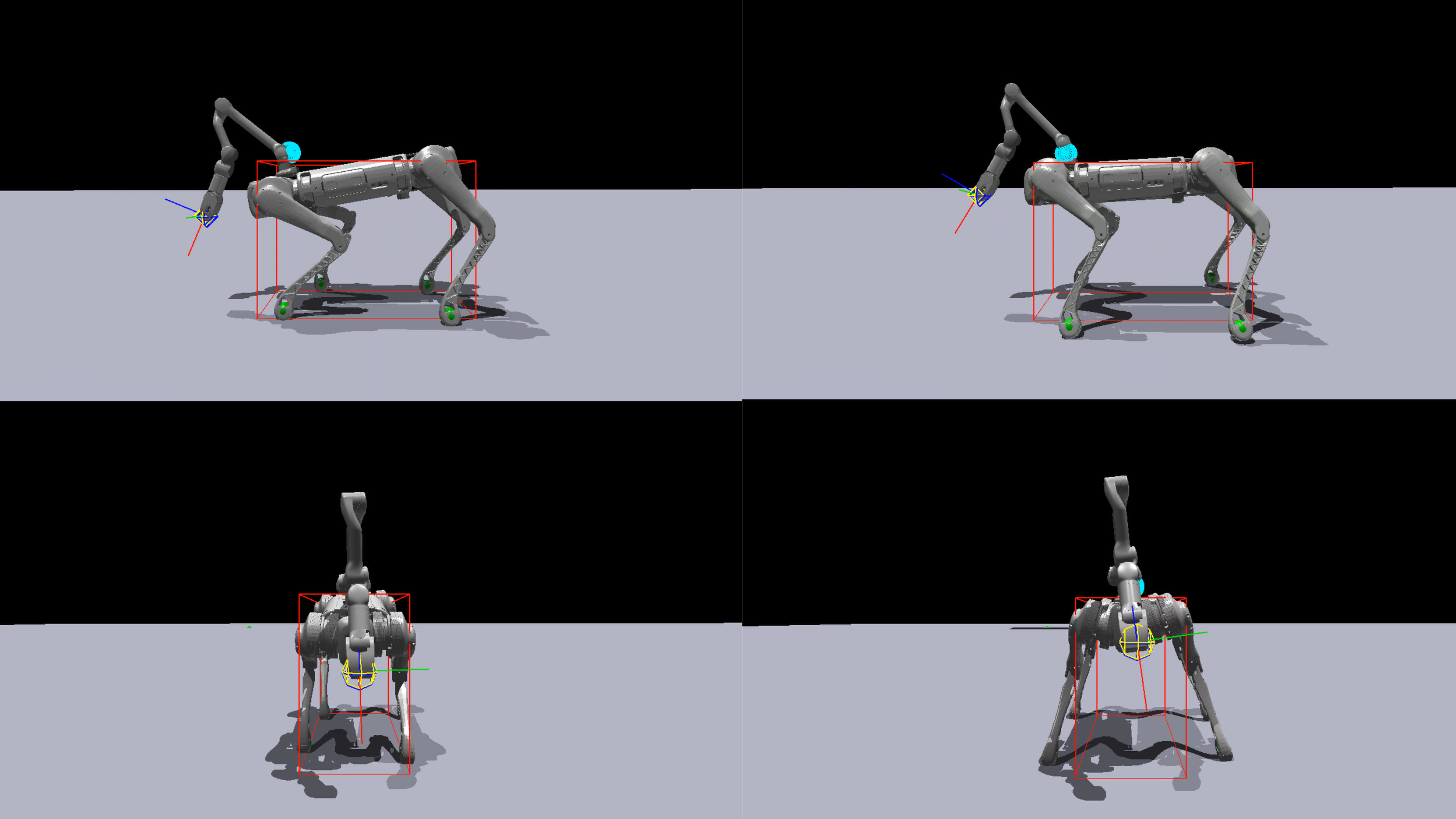}
    \caption{Snapshots in Isaac~Gym with low end-effector pose. Visible tilting and uneven support reflect exaggerated sensitivity to rigid corrective impulses.}
    \label{fig:isaac_lowpose_snapshots}%
    \vspace{-15pt}
\end{figure}

In all poses, the commanded velocities were broadly tracked, but systematic errors appeared (\cref{tab:sim2sim_rmse_summary}). Backward motions consistently yielded the largest translational errors, and the low EE pose produced elevated yaw noise. In the low configuration, forward tracking remained acceptable, but $v_x$ traces showed step-like artifacts, and the yaw exhibited persistent high-frequency oscillations (\cref{fig:base_velocity_tracking_isaac_low}). Visual snapshots (\cref{fig:isaac_lowpose_snapshots}) confirmed that lowering the arm shifted the center of mass and distorted the support geometry, producing asymmetric loading. Under PhysX’s rigid impulses, this translated into unstable ground reactions, explaining the degraded stability compared to mid and high poses.

\begin{table}[htbp!]
\centering
\vspace{-5pt}
\small
\renewcommand{\arraystretch}{1.15}
\begin{tabular}{lccc}
\toprule
\textbf{Motion type} & \textbf{Low EE} & \textbf{Mid EE} & \textbf{High EE} \\
\midrule
Straight (fwd/back) & 0.20 / 0.15 & 0.20 / 0.07 & 0.21 / 0.07 \\
Turning             & 0.08 / 0.27 & 0.08 / 0.21 & 0.09 / 0.21 \\
Combined (fwd+turn) & 0.17 / 0.23 & 0.18 / 0.18 & 0.18 / 0.18 \\
Neutral / Stop      & 0.06 / 0.30 & 0.04 / 0.05 & 0.04 / 0.05 \\
\bottomrule
\end{tabular}
\caption{Representative velocity tracking RMSE in \textbf{Isaac~Gym} across low, mid, and high end-effector poses. Errors are reported as $v_x$ / $v_{yaw}$ RMSE (m/s, rad/s). The low pose shows elevated yaw noise, especially in turning and combined motions.}
\label{tab:sim2sim_rmse_summary}
\vspace{-15pt}
\end{table}

\subsubsection{MuJoCo (default friction).} 
\begin{figure}[htbp]
    \centering
    \begin{subfigure}[t]{0.48\textwidth}
        \centering
        \includegraphics[width=\linewidth]{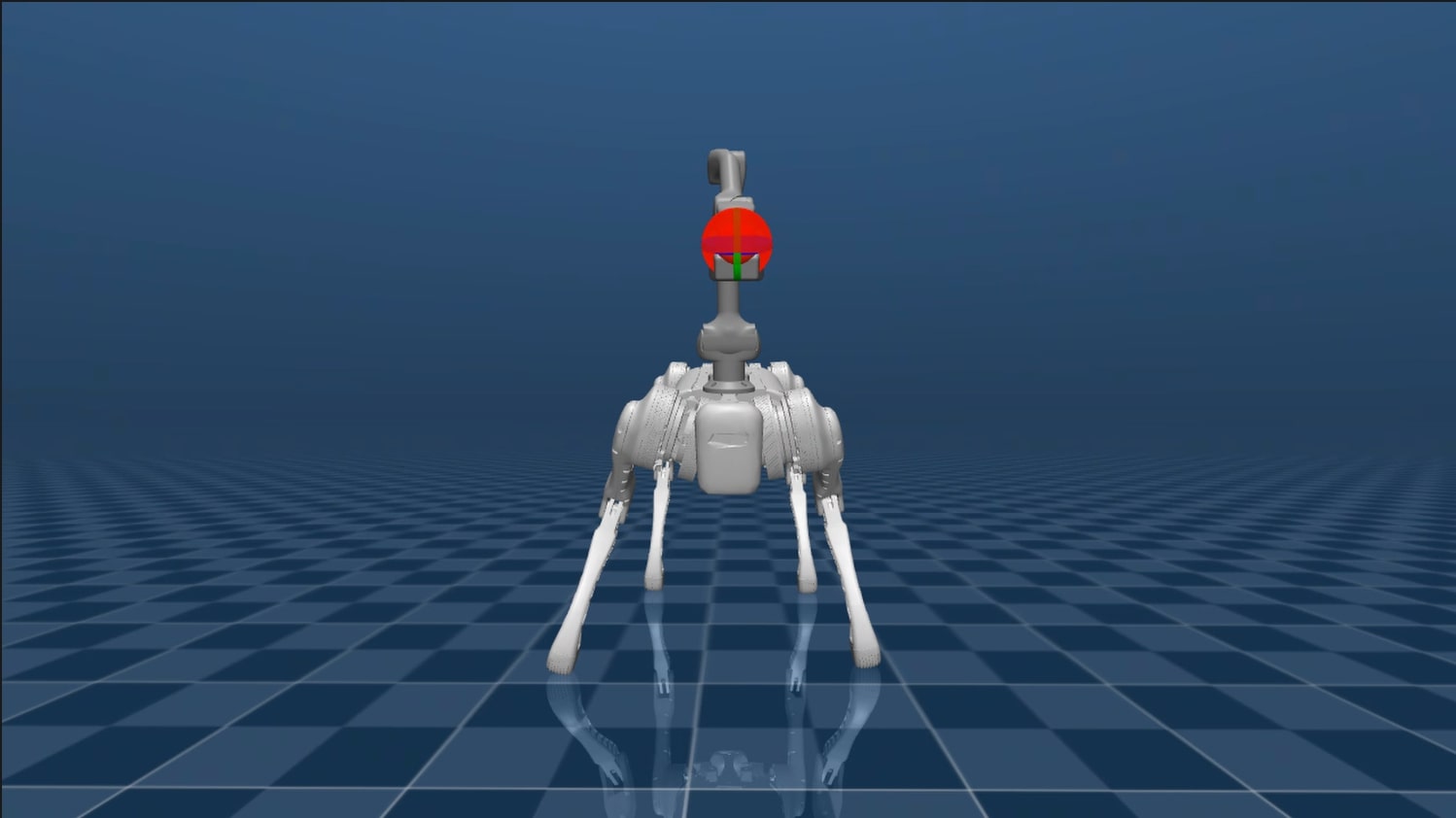}
    \end{subfigure}
    \begin{subfigure}[t]{0.48\textwidth}
        \centering
        \includegraphics[width=\linewidth]{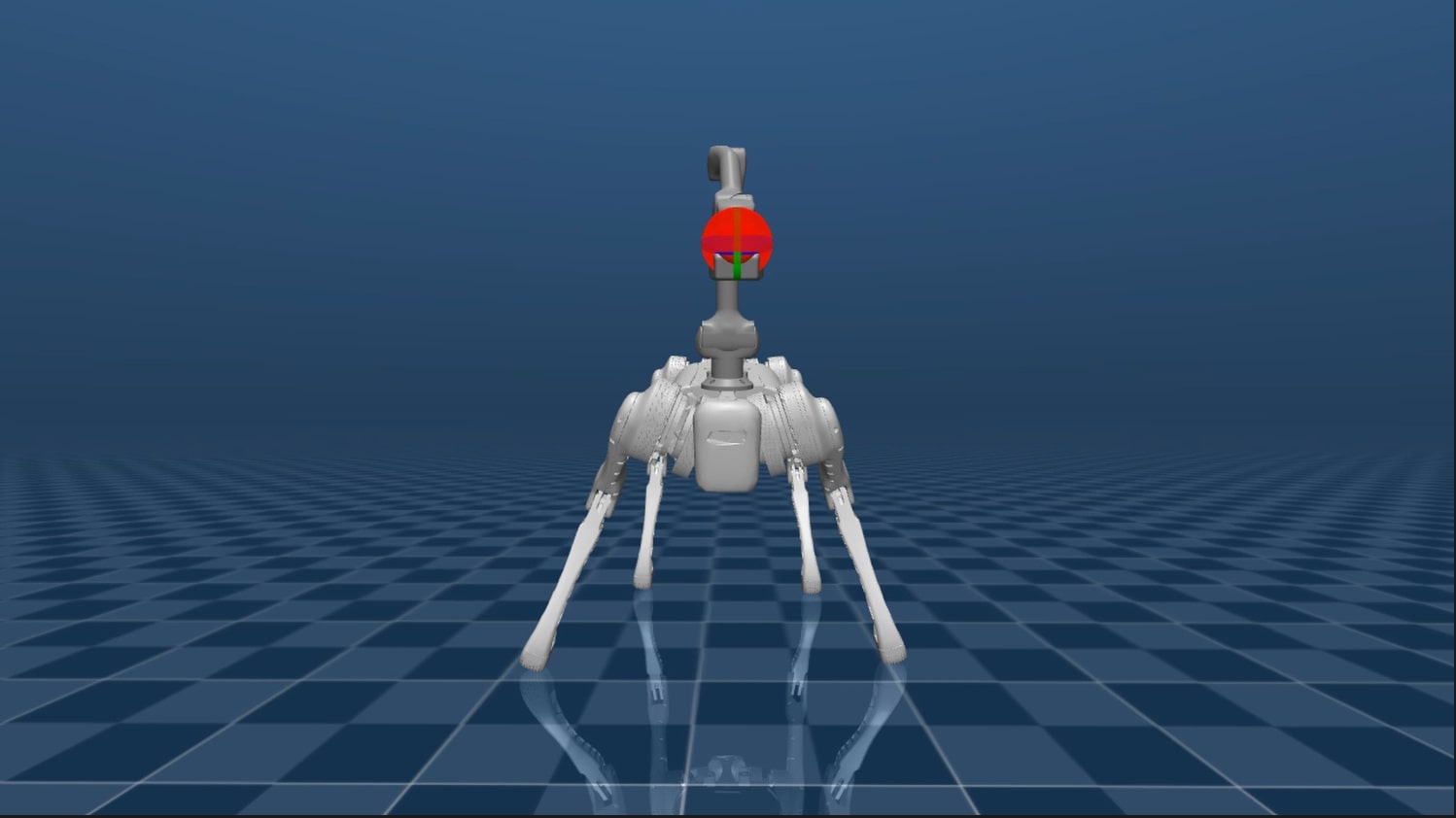}
    \end{subfigure}
    \caption{Foot sliding in MuJoCo under default \texttt{impratio=1} when standing still.}
    \label{fig:low_impratio_mujoco}
    \vspace{-5pt}
\end{figure}

In contrast, MuJoCo produced smoother dynamics with fewer oscillations and relatively uniform performance across poses. Even in the low configuration, both $v_x$ and yaw closely followed commands without instability. However, sliding artifacts were visible (\cref{fig:low_impratio_mujoco}), reflecting MuJoCo’s softened constraint solver.

\subsubsection{MuJoCo (tuned friction).}  
When evaluated with artificially enlarged impulse ratio (\texttt{impratio=100}), sliding was nearly eliminated and the neutral/stop phases reached near-zero error (\cref{tab:mujoco_tracking_rmse_summary}). This confirms that apparent smoothness reflects traction tuning rather than intrinsic robustness.

\begin{table}[htbp]
\centering
\small
\renewcommand{\arraystretch}{1.15}
\begin{tabular}{lccc}
\toprule
\textbf{Motion type} & \textbf{Low EE} & \textbf{Mid EE} & \textbf{High EE} \\
\midrule
Straight (fwd/back) & 0.15 / 0.08 & 0.14 / 0.07 & 0.12 / 0.08 \\
Turning             & 0.07 / 0.19 & 0.07 / 0.18 & 0.06 / 0.18 \\
Neutral / Stop      & 0.00 / 0.00 & 0.00 / 0.00 & 0.00 / 0.00 \\
\bottomrule
\end{tabular}
\caption{Velocity tracking RMSE in \textbf{MuJoCo} with tuned friction (\texttt{impratio=100}). 
Slipping is almost eliminated and tracking appears smoother, but this results from unrealistically high tangential impedance rather than true policy robustness. Values are $v_x$ / $v_{yaw}$ RMSE (m/s, rad/s).}
\label{tab:mujoco_tracking_rmse_summary}
\vspace{-20pt}
\end{table}

\subsubsection{Cross-simulator comparison.} 
The average results (\cref{tab:pose_avg_rmse}) highlight two consistent findings. First, the low EE pose is most error-prone, while the high pose yields the lowest RMSE, independent of the simulator. Second, absolute error levels depend strongly on contact modeling: Isaac~Gym introduces yaw oscillations, whereas MuJoCo exhibits foot drift. Tuning MuJoCo friction reduces drift further, but reflects unrealistic traction rather than true policy robustness.

\begin{table}[htbp]
\centering
\begin{tabular}{lccc}
\toprule
\textbf{Simulator} & \textbf{Low EE (avg)} & \textbf{Mid EE (avg)} & \textbf{High EE (avg)} \\
\midrule
Isaac Gym            & 0.144 / 0.231 & 0.144 / 0.118 & 0.163 / 0.114 \\
MuJoCo (default)     & 0.113 / 0.123 & 0.123 / 0.118 & 0.123 / 0.122 \\
MuJoCo (\texttt{impratio=100}) & 0.118 / 0.120 & 0.109 / 0.110 & 0.112 / 0.118 \\
\bottomrule
\end{tabular}
\caption{Average velocity tracking RMSE ($v_x$ / $v_{yaw}$) across motion patterns for low, mid, and high EE poses. Isaac Gym exhibits yaw oscillations, MuJoCo drift, and tuned MuJoCo shows artefactually low error.}
\label{tab:pose_avg_rmse}
\vspace{-15pt}
\end{table}

\subsection{Sim-to-real tracking}
On the physical B1+Z1 platform, the deployed policy successfully executed locomotion across all tested end-effector (EE) poses, consistent with simulation trends. The \textbf{mid-pose} configurations produced the most stable behaviour, with smooth trajectories and minimal oscillations even at maximum teleoperation speeds (0.4,m/s, 0.8,rad/s).

In contrast, the \textbf{high-pose} setting induced base shaking, suggesting overcompensation for ground contacts—similar to oscillations observed in Isaac~Gym during sim-to-sim evaluation. The \textbf{low-pose} configuration maintained stability but showed reduced gait fluidity and slower forward speed due to tilted base geometry and reduced ground clearance. Unlike simulation, however, persistent oscillations were not observed, reflecting real–sim dynamics differences.

\begin{figure}[htbp]
    \centering
    \vspace{-15pt}
    \includegraphics[width=0.7\linewidth]{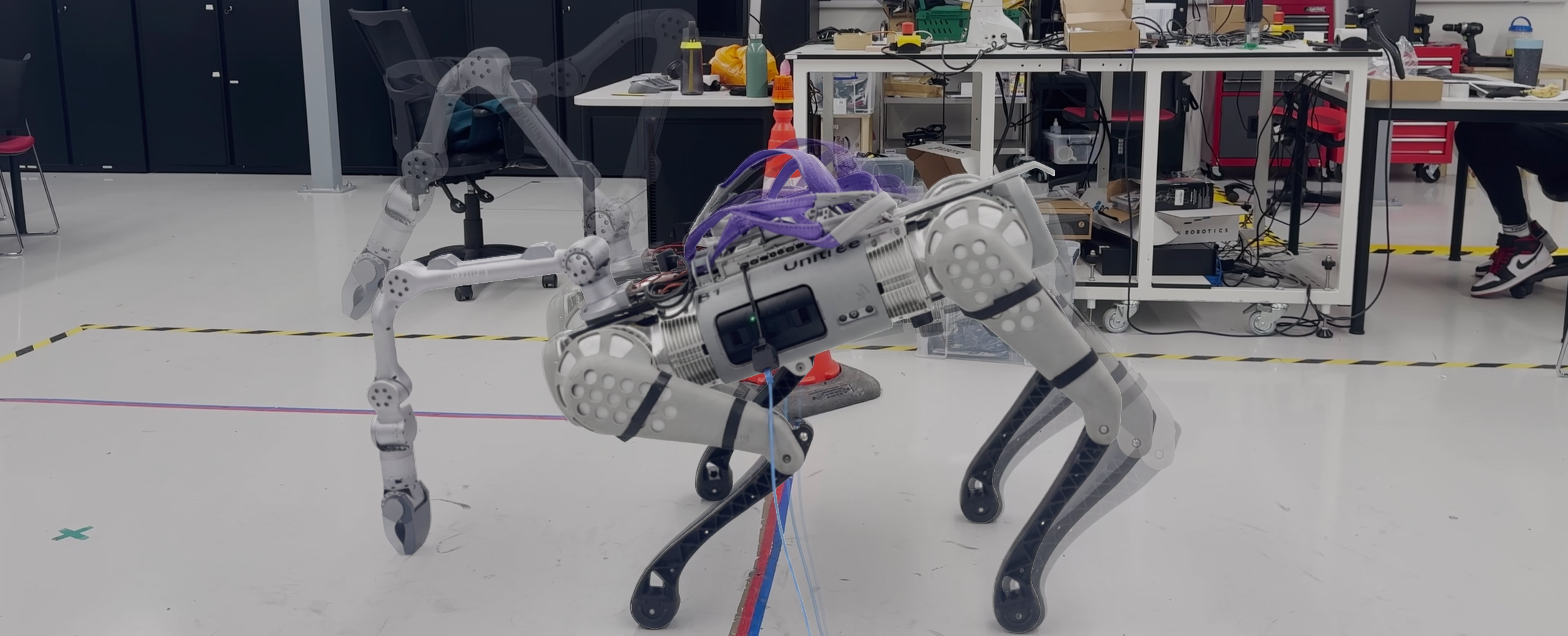}
    \caption{Reaching range test}
    \label{fig:reaching_test_vid}
    \vspace{-15pt}
\end{figure}

Fig.~\ref{fig:reaching_test_vid} illustrates a separate \textbf{reaching range} test, where manual arm adjustments confirmed that the controller adapts the trunk posture to extend manipulation reach while preserving stability.


Overall, these real-world results confirm that the policy generalises across EE configurations. Mid poses provide the best trade-off between manipulation reach and locomotion stability, while extreme poses expose controller limits consistent with sim-to-sim findings.



\subsection{Realworld Teleoperation picking}

\begin{figure}[htbp]
    \vspace{-15pt}
    \centering
    \includegraphics[width=0.9\linewidth]{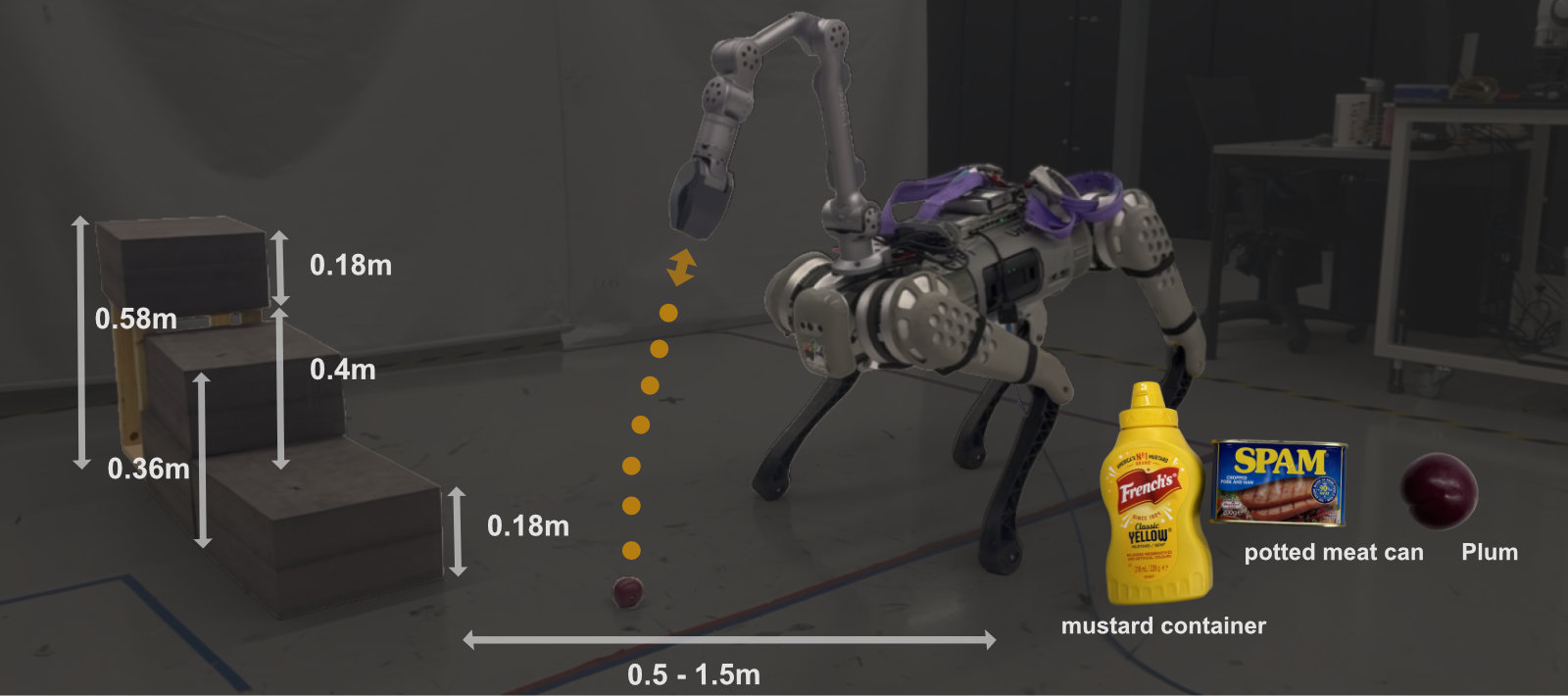}
    \caption{Teleoperated object-picking setup. \textbf{Bottom right}: Selected YCB objects. \textbf{Left}: Experimental arrangement with objects placed at heights of 0.18–0.58\,m and distances of 0.5–1.5\,m from the robot base.}
    \label{fig:object_picking_setup}%
    \vspace{-15pt}
\end{figure}

To assess the real-world performance of our deployed policy, we conducted teleoperated object-picking experiments using three representative YCB objects~\cite{calli2015benchmarking}: a mustard container (tall bottle-like geometry), a potted meat can (compact box-like object), and a plum (spherical, difficult to grasp).  Objects were placed at varying heights (0–0.58,m) and distances (0.5–1.5,m) relative to the robot, as shown in Fig.~\ref{fig:object_picking_setup}.




\begin{figure}[htbp]
    \vspace{-15pt}
    \centering
    \subfloat[Baseline]{%
        \includegraphics[width=0.49\columnwidth]{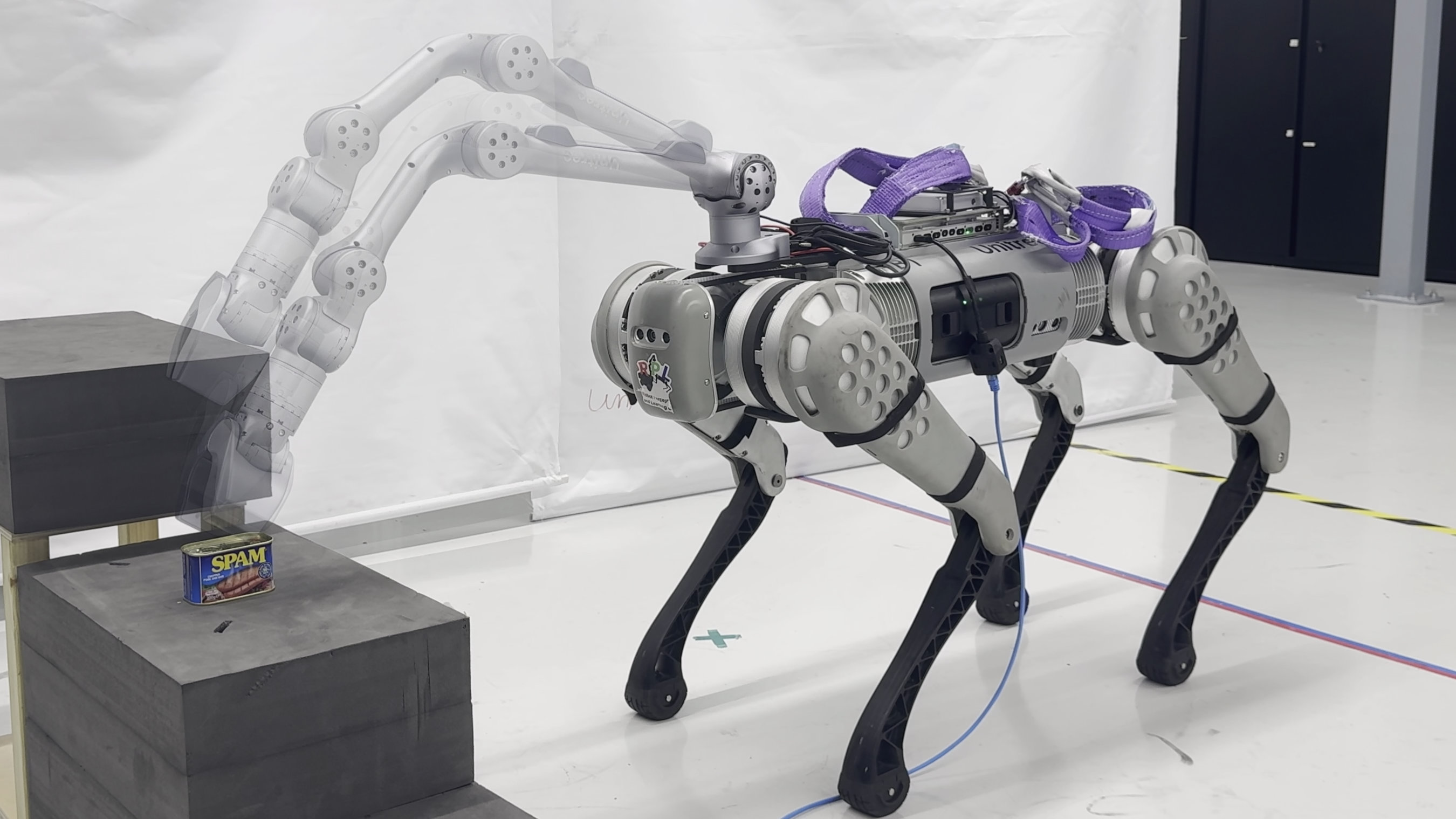}
    } 
    \subfloat[Whole-Body Control]{%
        \includegraphics[width=0.49\columnwidth]{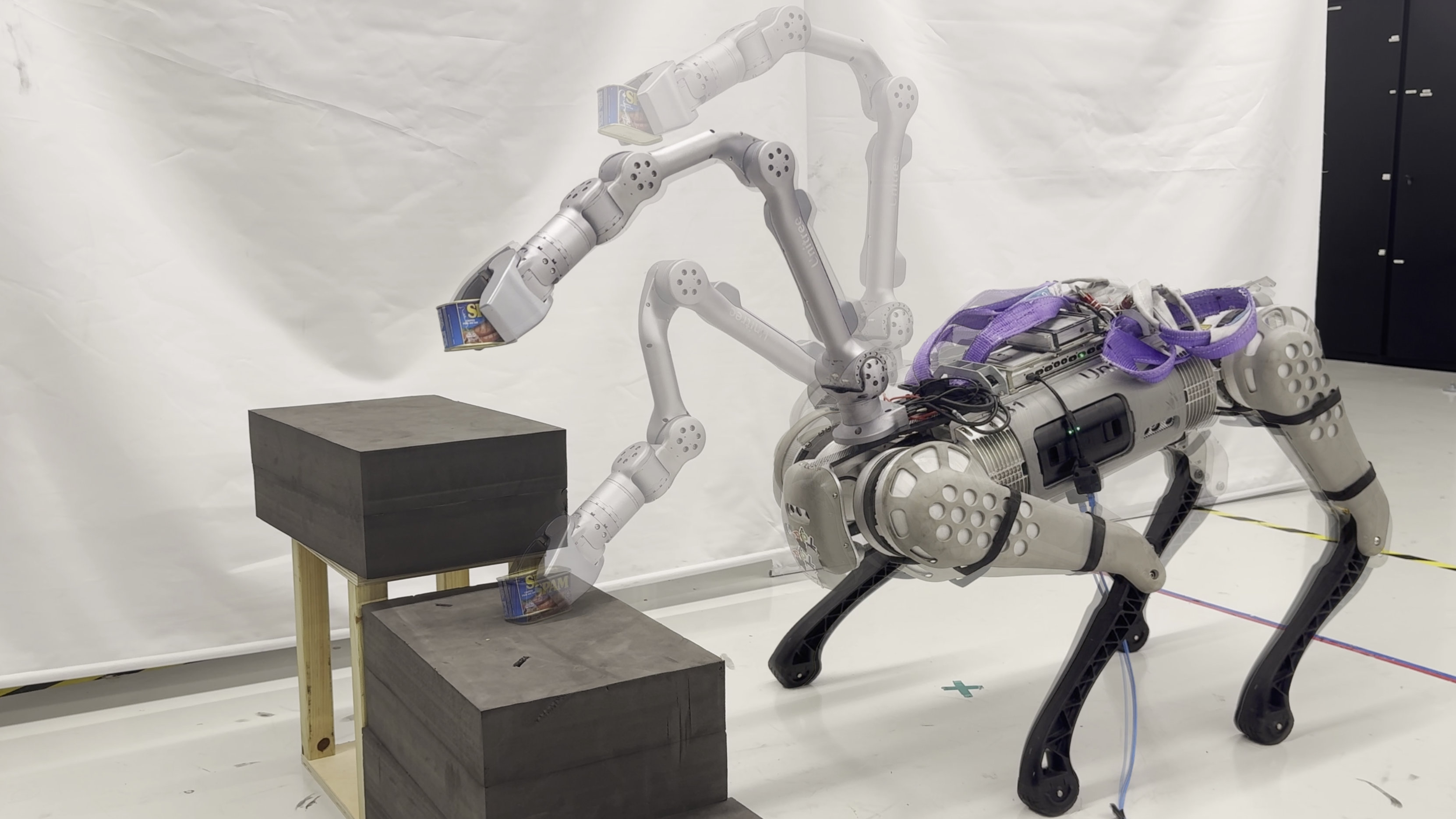}
    }
    \caption{Object picking under two strategies. 
    \textbf{Left}: floating-base arm control at fixed body height. 
    \textbf{Right}: Whole-Body Control (WBC) coordinating arm and leg motions for extended reach and stability.}
    \label{fig:floating_base_vs_wbc}
    \vspace{-15pt}
\end{figure}

We compared two teleoperation modes: 1) \textbf{Baseline}: Unitree’s default controller, with separate arm–base control. 2) \textbf{WBC}: Our unified Whole-Body Controller coordinating arm and base. As illustrated in Fig.~\ref{fig:floating_base_vs_wbc}, baseline teleoperation was limited by fixed trunk posture, while WBC adapted stance and orientation to extend reach and improve stability. Each object was tested three times per height.



\begin{table}[htbp]
\centering
\scriptsize
\renewcommand{\arraystretch}{1.25}
\begin{tabular}{lcccc}
\toprule
\textbf{Object} & \textbf{Height (m)} & \textbf{Trials} & \textbf{Floating Base (success)} & \textbf{WBC (success)} \\
\midrule
Potted meat can & 0.58 & 3 & 3/3 & 3/3 \\
                & 0.36 & 3 & 1/3 (reach limit) & 2/3 \\
                & 0.18 & 3 & 0/3 (reach limit) & 3/3 \\
                & 0.00 & 1 & 0/1 (not reachable) & 1/1 \\
\midrule
Mustard container & 0.58 & 3 & 3/3 & 3/3 \\
                  & 0.36 & 3 & 2/3 & 3/3 \\
                  & 0.18 & 3 & 0/3 (reach limit) & 3/3 \\
                  & 0.00 & 1 & 0/1 (not reachable) & 1/1 \\
\midrule
Plum (sphere) & 0.58 & 3 & 2/3 (slippery) & 2/3 (slippery) \\
              & 0.36 & 3 & 0/3 (reach/workspace limit) & 2/3 \\
              & 0.18 & 3 & 0/3 (reach/workspace limit) & 2/3 (slip once) \\
              & 0.00 & 1 & 0/1 (not reachable) & 1/1 \\
\bottomrule
\end{tabular}
\caption{Teleoperated pick success rate across control modes.}
\label{tab:teleop_results}
\vspace{-15pt}
\end{table}

Results (Tab.~\ref{tab:teleop_results}) show that both controllers performed similarly for high objects (0.58\,m), but baseline control failed consistently at low heights (0.18\,m and ground) due to reach limitations. WBC achieved near-perfect success across all heights by adapting body posture, though spherical objects like the plum remained challenging due to gripper constraints.
\vspace{-5pt}

\section{Conclusion}\label{sec:conclusion}
This work addressed the deployment gap in quadruped loco-manipulation by emphasizing reproducibility and transfer rather than algorithmic novelty. We investigated how RL-based whole-body controllers, trained in large-scale simulation, transfer across simulators and to real hardware. Cross-simulator studies revealed consistent behavior but exposed sensitivities to contact modeling: Isaac~Gym introduced oscillatory artifacts, while MuJoCo produced smoother dynamics but with foot drift under default friction. Real-world deployment on Unitree B1+Z1 confirmed that policies trained in simulation generalize, with mid-pose configurations yielding the most stable locomotion. However, extreme poses exposed limits in compensation strategies, consistent with sim-to-sim findings. Teleoperated object-picking further showed that whole-body coordination extends reachability and improves stability compared to floating-base baselines, although grasp success remained constrained by object geometry and hardware. Finally, we contribute a reproducible deployment pipeline that supports sim-to-sim verification and real-robot testing. Our results highlight contact modeling and robustness as central bottlenecks.

\vspace{-10pt}
\begin{credits}
\subsubsection{\ackname} This work was partially supported by UKRI FLF [MR/V025333/1] (RoboHike). For Open Access, the author has applied a CC BY copyright license to any manuscript version arising from this submission.
\end{credits}
\vspace{-10pt}
%
%
\bibliographystyle{splncs04}
\bibliography{references}
\end{document}